\documentclass[11pt]{article}

\usepackage[preprint]{acl}
\usepackage{subcaption}

\usepackage{tabularx} 
\usepackage{booktabs}
\usepackage{multirow}
\usepackage{amsmath}
\usepackage{amssymb}
\usepackage{url}
\usepackage{xcolor}
\usepackage{times}
\usepackage{latexsym}

\usepackage[T1]{fontenc}
\usepackage{enumitem}
\usepackage[utf8]{inputenc}

\usepackage{microtype}

\usepackage{inconsolata}

\usepackage{graphicx}
\definecolor{variantgray}{gray}{0.55}

\title{How Closely Do LLM Reviews Align with Human Peer Review?}

\author{Abraham Camelo-Guerrero \\
School of Information Technology\\
  York University\\
  Toronto, Ontario M3J 1P3\\
  \texttt{acamelog@yorku.ca} \\\And
  Jairo Diaz-Rodriguez \\
Department of Mathematics and Statistics\\
  York University\\
  Toronto, Ontario M3J 1P3\\
  \texttt{jdiazrod@yorku.ca} \\}

\begin{document}
\maketitle
\begin{abstract}

Large language models (LLMs) are increasingly used to generate scientific reviews, yet existing evaluations rarely examine whether different providers align with both conference decisions and human reviewing priorities within the same controlled setting. We compare reviews from OpenAI GPT-5.4, Google Gemini 3.1 Pro Preview, and Anthropic Claude Opus 4.6 with human reviews and final decisions for 300 topic-matched ICLR 2026 submissions, equally divided among oral, poster, and rejected papers. Each model reviewed every paper using identical instructions and rating scales after decision information was removed. Our study contributes a cross-provider analysis of three complementary dimensions: alignment with broad and fine-grained decision categories, differences in recommendation-scale usage, and thematic agreement in identified weaknesses. All three LLMs distinguished accepted from rejected papers, but none reproduced the oral versus poster distinction present in human ratings. Scoring patterns were provider-specific: Gemini assigned systematically higher ratings, while OpenAI and Claude were closer to humans for rejected and poster papers but more critical of oral papers. Human and LLM reviews also differed in emphasis, with LLMs more frequently identifying missing baseline comparisons and humans more often raising computational-efficiency concerns. These results show that broad decision alignment does not imply agreement with finer human judgments or reviewing priorities.

\end{abstract}

\section{Introduction}
\label{sec:introduction}

Peer review is a central mechanism for evaluating the validity, relevance, and contribution of scientific research. Conference reviews in machine learning and natural language processing typically combine written assessments with numerical recommendations that inform acceptance decisions and presentation categories. Recent advances in large language models (LLMs) have made it possible to automate parts of this process. General-purpose and specialized models can summarize contributions, identify strengths and weaknesses, formulate questions for authors, and assign numerical recommendations. However, producing a fluent and appropriately structured review does not establish that the model's judgment aligns with expert peer review.

Evaluating LLM-generated reviews therefore requires examining several complementary forms of alignment. First, an automated reviewer should exhibit \textit{outcome alignment}, assigning systematically different ratings to papers associated with different conference outcomes. Second, its use of the recommendation scale should be compared with that of human reviewers and other models. Third, its \textit{review-content alignment} should be assessed by examining whether it identifies and prioritizes similar weaknesses. Existing work has investigated individual aspects of automated reviewing, LLM-based evaluation, and review-text quality, but these signals have rarely been examined together in a controlled cross-provider setting.

This study compares reviews generated by three contemporary LLMs with human reviews and final decisions from the International Conference on Learning Representations (ICLR). We analyze 300 ICLR 2026 submissions, equally divided among oral, poster, and rejected papers. To reduce topical differences among decision categories, oral papers were matched with thematically similar poster and rejected papers using abstract embeddings and one-to-one similarity matching. Decision information displayed in the PDFs was removed before review generation, and each paper was independently evaluated by models from OpenAI, Google, and Anthropic using the same instructions and rating scale. We compare rating distributions across evaluators and decision categories and analyze the weaknesses identified by human and LLM reviewers through embedding-based thematic clustering.

The results reveal a distinction between broad outcome discrimination and finer alignment with human judgments. All three LLMs assigned higher ratings to accepted papers than to rejected papers, indicating that they captured a broad acceptance-related signal. However, none reproduced the distinction between poster and oral papers observed in the human ratings. Scoring patterns also varied across providers. Gemini generally assigned more favorable ratings, while OpenAI and Claude were closer to human reviewers for rejected and poster papers but more critical when evaluating oral papers. The models also used a narrower portion of the rating scale. Finally, human and LLM reviews shared several major weakness themes but differed in emphasis: LLMs more frequently criticized insufficient baseline comparisons, whereas human reviewers more often focused on computational efficiency and resource requirements.

This work makes three contributions:

\begin{enumerate}[leftmargin=*]
\item We provide a controlled cross-provider evaluation of LLM-generated peer-review scores across broad acceptance outcomes and finer presentation categories.

\item We show that distinguishing accepted from rejected papers does not imply alignment with finer conference decisions or equivalent use of the recommendation scale.

\item We compare the thematic priorities of human and LLM criticisms, demonstrating that disagreement extends beyond numerical ratings to the weaknesses considered salient.

\end{enumerate}

Together, these findings support the use of LLMs as sources of supplementary critique while cautioning against treating their scores and priorities as interchangeable with human peer-review judgments.

\section{Related Work}
\label{sec:related_work}

\subsection{Computational Analysis of Peer Review}

Peer review is not a perfectly consistent measurement process. Reviewers evaluating the same manuscript may emphasize different aspects of the work, identify different weaknesses, or reach conflicting recommendations \cite{kumar2023reviewers,malicki2024structured}. This disagreement does not make peer review uninformative, but it complicates the evaluation of automated reviewing systems. Rather than assuming that each paper has a single objectively correct score, automated reviewers should be compared with the variation and structure of actual reviewing behavior.

The release of datasets such as PeerRead has enabled peer-review judgments to be studied computationally by connecting paper content, written reviews, numerical scores, and acceptance decisions \cite{kang2018peerread}. Such resources support analyses of reviewer disagreement, score distributions, review content, and the relationship between written assessments and publication outcomes. Our work follows this line of research but focuses specifically on comparing contemporary LLM-generated reviews with human ratings and final conference decisions.

\subsection{LLMs as Evaluators}

The evaluation of scientific papers is related to the broader use of LLMs as judges. LLM-based evaluation has achieved promising agreement with human judgments in tasks such as natural-language generation assessment \cite{chiang2023llmeval,liu2023geval}. However, large-scale studies show that this agreement varies substantially across models, datasets, evaluation properties, and prompting conditions \cite{bavaresco2025judges}. LLM evaluators can also exhibit systematic biases in how they assign preferences and numerical scores \cite{wang2024fairevaluators}.

These findings indicate that evaluator quality cannot be established solely through aggregate agreement with human labels. Models may differ in scale usage, severity, sensitivity to particular evaluation criteria, and stability across settings. Our analysis applies this perspective to scientific peer review by separately examining outcome alignment, rating-distribution differences, and the themes emphasized in written criticisms.

\subsection{Automated Scientific Reviewing}

Recent work has explored the use of LLMs to generate structured scientific reviews. Systems such as OpenReviewer and DeepReview investigate domain-specific training, structured reasoning, literature retrieval, and evidence-based argumentation for automated paper assessment \cite{idahl2025openreviewer,zhu2025deepreview}. Related research has developed methods for evaluating the quality of review text, including whether criticisms are supported by evidence from the paper \cite{guo2023substantiation}. These systems suggest that LLMs may assist authors seeking preliminary feedback or help reviewers organize their assessments.

At the same time, research on automated scientific reviewing has identified important limitations. CLAIMCHECK shows that producing criticisms that are sound and grounded in a paper's claims remains difficult, even for advanced models \cite{ou2025claimcheck}. PeerCheck identifies differences between the terms and aspects emphasized by LLM-generated and human-written reviews and finds that interventions intended to increase human similarity do not improve every model consistently \cite{chen2026peercheck}. Reviewer instructions also affect automated judgments: official conference guidelines can improve alignment with human evaluations, whereas rigid rubric-based scoring may reduce it \cite{li2026guidelines}.

Collectively, these studies demonstrate that automated-review quality depends on both the underlying model and the evaluation procedure. However, the relationship among three signals remains insufficiently characterized within a single controlled setting: whether frontier models distinguish broad and fine-grained conference outcomes, whether different providers use the recommendation scale similarly, and whether their written reviews prioritize the same weaknesses as human reviewers. Our study addresses this gap through a cross-provider comparison using topic-matched ICLR submissions, human reviews, and final conference decisions.

\begin{table*}[t]
\centering
\caption{Metadata collected for each review.}
\label{tab:review_metadata}

\scriptsize
\setlength{\tabcolsep}{4pt}
\renewcommand{\arraystretch}{1.05}

\begin{tabularx}{\textwidth}{
    >{\raggedright\arraybackslash}p{0.17\textwidth}
    >{\raggedright\arraybackslash}X
}
\toprule
\textbf{Field} & \textbf{Definition for ICLR reviews} \\
\midrule

\texttt{paper\_id}
&
Unique OpenReview identifier of the ICLR submission associated with the review.
\\

\texttt{paper\_number}
&
Numerical submission number assigned to the paper within that year's ICLR conference.
\\

\texttt{paper\_title}
&
Title of the ICLR submission evaluated by the reviewer.
\\

\texttt{decision}
&
Final decision assigned to the submission, such as Oral, Spotlight, Poster, Reject, Withdrawn, or Desk Reject. The available categories may vary by ICLR year.
\\

\texttt{review\_id}
&
Unique OpenReview identifier assigned to the individual official review.
\\

\texttt{reviewer}
&
Anonymous OpenReview identifier of the reviewer who submitted the review. It does not normally reveal the reviewer's identity.
\\

\texttt{rating}
&
Reviewer's overall recommendation regarding whether the paper should be accepted or rejected. The rating reflects the paper's overall value to the ICLR community and is supported by the reviewer's written assessment.
\\

\texttt{confidence}
&
Reviewer's self-assessment of confidence in the evaluation, considering familiarity with the topic and certainty about the review.
\\

\texttt{soundness}
&
Reviewer's assessment of whether the paper's claims are adequately supported and whether its theoretical or empirical results are technically correct and scientifically rigorous.
\\

\texttt{presentation}
&
Reviewer's assessment of how clearly and effectively the paper communicates its ideas, methods, arguments, and results.
\\

\texttt{contribution}
&
Reviewer's assessment of whether the paper provides new, relevant, and sufficiently valuable knowledge to the ICLR community. This does not require state-of-the-art performance.
\\

\texttt{summary}
&
Reviewer's summary of the paper's main objective, approach, claims, findings, and intended contributions.
\\

\texttt{strengths}
&
Strong aspects identified by the reviewer, such as technical correctness, experimental rigor, reproducibility, clarity, novelty, relevance, or significance.
\\

\texttt{weaknesses}
&
Weak aspects identified by the reviewer, particularly concerns related to correctness, rigor, reproducibility, clarity, novelty, relevance, or support for the paper's claims.
\\

\texttt{questions}
&
Questions directed to the authors to clarify the reviewer's understanding or obtain evidence needed to evaluate the paper more confidently.
\\

\bottomrule
\end{tabularx}
\end{table*}

\section{Methodology}
\label{sec:methodology}


\begin{figure*}[t]
    \centering
    \includegraphics[width=0.24\textwidth]{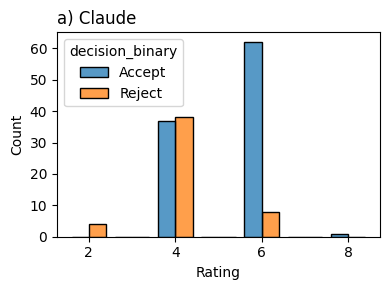}\hfill
    \includegraphics[width=0.24\textwidth]{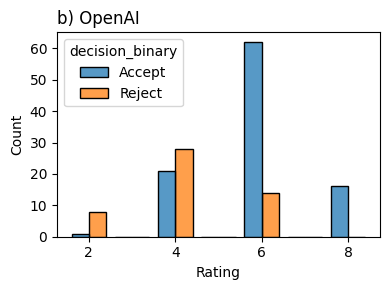}\hfill
    \includegraphics[width=0.24\textwidth]{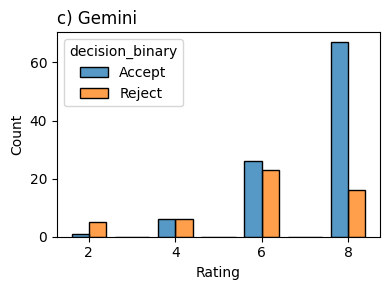}\hfill
    \includegraphics[width=0.24\textwidth]{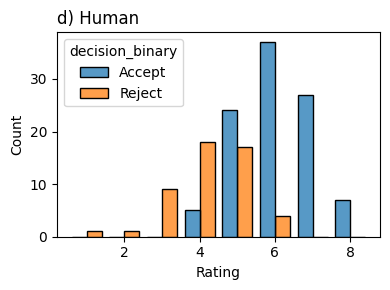}
    
    \caption{Rating distributions for accepted and rejected papers according to (a) Claude, (b) OpenAI, (c) Gemini, and (d) human reviewers. Oral and poster papers were combined into the accepted category.}
    \label{fig:accepted_rejected_distributions}
\end{figure*}

\begin{table*}[t]
\centering
\caption{Mann--Whitney \(U\) comparisons between ratings of accepted and rejected papers.}
\label{tab:accepted_rejected_comparison}

\small
\setlength{\tabcolsep}{10pt}
\renewcommand{\arraystretch}{1.15}

\begin{tabular}{@{}lcccc@{}}
\toprule
\textbf{Statistic} &
\textbf{Claude} &
\textbf{OpenAI} &
\textbf{Gemini} &
\textbf{Human} \\
\midrule

Accepted, \(n\) papers
& 200
& 200
& 200
& 200 \\

Rejected, \(n\) papers
& 100
& 100
& 100
& 100 \\

Accepted median
& 6.0
& 6.0
& 8.0
& 6.0 \\

Rejected median
& 4.0
& 4.0
& 6.0
& 4.0 \\

Mann--Whitney \(U\) statistic
& 3753.0
& 3932.0
& 3449.5
& 4580.0 \\

\(p\)-value
& \(1.36 \times 10^{-8}\)
& \(4.09 \times 10^{-10}\)
& \(2.21 \times 10^{-5}\)
& \(8.30 \times 10^{-17}\) \\

\bottomrule
\end{tabular}
\end{table*}
\begin{figure*}[t]
    \centering
    \includegraphics[width=0.24\textwidth]{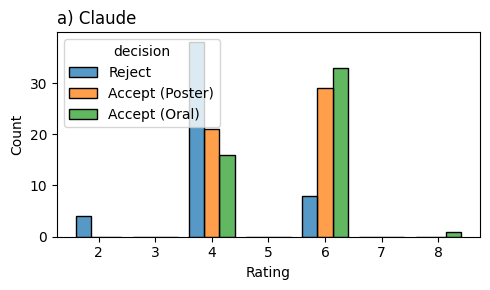}\hfill
    \includegraphics[width=0.24\textwidth]{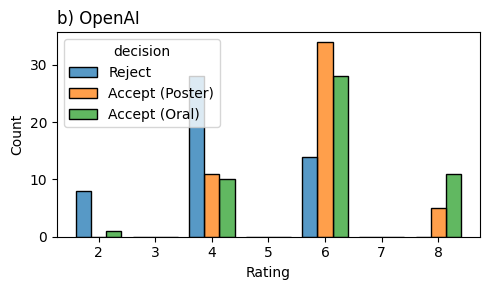}\hfill
    \includegraphics[width=0.24\textwidth]{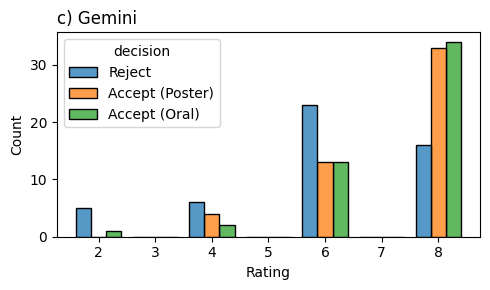}\hfill
    \includegraphics[width=0.24\textwidth]{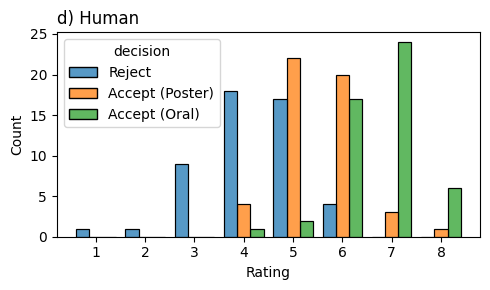}
    
    \caption{Rating distributions across the three ICLR decision categories for (a) Claude, (b) OpenAI, (c) Gemini, and (d) human reviewers.}
    \label{fig:decision_category_distributions}
\end{figure*}

\begin{table*}[t]
\centering
\caption{Pairwise comparisons of rating distributions across decision categories by evaluator.}
\label{tab:decision_category_comparisons}

\scriptsize
\setlength{\tabcolsep}{3.5pt}
\renewcommand{\arraystretch}{1.15}

\resizebox{\textwidth}{!}{%
\begin{tabular}{
    @{}
    ll
    ccc
    ccc
    ccc
    ccc
    @{}
}
\toprule

\multirow{2}{*}{\shortstack{\textbf{Group}\\\textbf{A}}} &
\multirow{2}{*}{\shortstack{\textbf{Group}\\\textbf{B}}} &
\multicolumn{3}{c}{\textbf{Claude}} &
\multicolumn{3}{c}{\textbf{OpenAI}} &
\multicolumn{3}{c}{\textbf{Gemini}} &
\multicolumn{3}{c}{\textbf{Human}} \\

\cmidrule(lr){3-5}
\cmidrule(lr){6-8}
\cmidrule(lr){9-11}
\cmidrule(l){12-14}

& &
\shortstack{\textbf{Median}\\\textbf{A}} &
\shortstack{\textbf{Median}\\\textbf{B}} &
\shortstack{\textbf{Holm-adjusted}\\\(\boldsymbol{p}\)} &
\shortstack{\textbf{Median}\\\textbf{A}} &
\shortstack{\textbf{Median}\\\textbf{B}} &
\shortstack{\textbf{Holm-adjusted}\\\(\boldsymbol{p}\)} &
\shortstack{\textbf{Median}\\\textbf{A}} &
\shortstack{\textbf{Median}\\\textbf{B}} &
\shortstack{\textbf{Holm-adjusted}\\\(\boldsymbol{p}\)} &
\shortstack{\textbf{Median}\\\textbf{A}} &
\shortstack{\textbf{Median}\\\textbf{B}} &
\shortstack{\textbf{Holm-adjusted}\\\(\boldsymbol{p}\)} \\

\midrule

Reject &
Oral &
4.0 & 6.0 & \(2.33 \times 10^{-7}\) &
4.0 & 6.0 & \(2.24 \times 10^{-7}\) &
6.0 & 8.0 & \(6.43 \times 10^{-4}\) &
4.0 & 6.5 & \(4.23 \times 10^{-16}\) \\

Reject &
Poster &
4.0 & 6.0 & \(1.19 \times 10^{-5}\) &
4.0 & 6.0 & \(2.24 \times 10^{-7}\) &
6.0 & 8.0 & \(1.04 \times 10^{-3}\) &
4.0 & 5.33 & \(1.59 \times 10^{-9}\) \\

Poster &
Oral &
6.0 & 6.0 & \(0.256\) &
6.0 & 6.0 & \(0.376\) &
8.0 & 8.0 & \(0.809\) &
5.33 & 6.5 & \(3.24 \times 10^{-9}\) \\

\bottomrule
\end{tabular}%
}
\end{table*}

\begin{figure*}[t]
    \centering
    \includegraphics[width=0.75\textwidth]{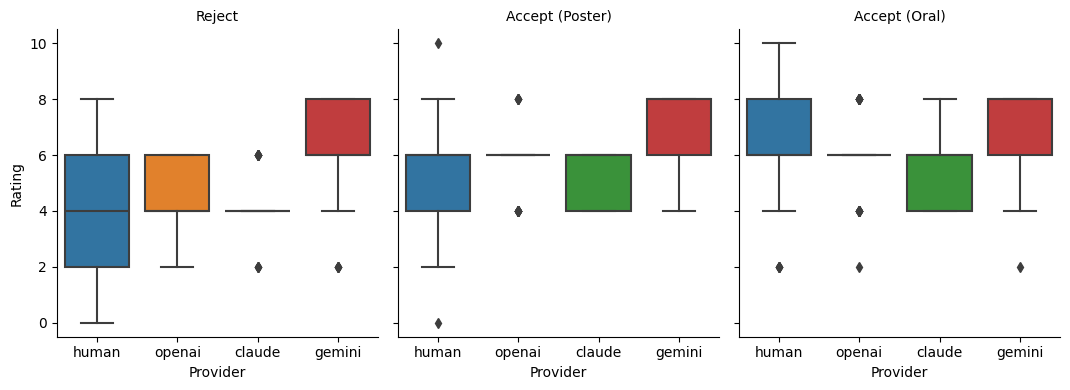}
    \caption{Rating distributions for human reviewers and the three LLM providers across rejected, poster, and oral papers. Boxes represent the interquartile range, horizontal lines indicate the median, whiskers show the non-outlier range, and individual markers represent outliers.}
    \label{fig:llm_rating_comparison}
\end{figure*}

\begin{table*}[t]
\centering
\caption{Paired Wilcoxon signed-rank comparisons among LLM evaluators within each decision category. Wilcoxon \(W\) statistics and Holm-adjusted \(p\)-values are reported.}
\label{tab:llm_pairwise_comparisons}

\small
\setlength{\tabcolsep}{7pt}
\renewcommand{\arraystretch}{1.12}

\begin{tabular}{@{}llcccccc@{}}
\toprule
\multirow{2}{*}{\textbf{Provider A}} &
\multirow{2}{*}{\textbf{Provider B}} &
\multicolumn{2}{c}{\textbf{Reject}} &
\multicolumn{2}{c}{\textbf{Accept (Poster)}} &
\multicolumn{2}{c}{\textbf{Accept (Oral)}} \\
\cmidrule(lr){3-4}
\cmidrule(lr){5-6}
\cmidrule(lr){7-8}

& &
\(\boldsymbol{W}\) &
\textbf{Adjusted \(p\)} &
\(\boldsymbol{W}\) &
\textbf{Adjusted \(p\)} &
\(\boldsymbol{W}\) &
\textbf{Adjusted \(p\)} \\
\midrule

OpenAI
& Claude
& 94.5
& \(1.000\)
& 44.0
& \(8.10 \times 10^{-3}\)
& 87.5
& \(2.65 \times 10^{-2}\) \\

OpenAI
& Gemini
& 43.5
& \(1.50 \times 10^{-6}\)
& 15.0
& \(8.71 \times 10^{-7}\)
& 29.5
& \(1.23 \times 10^{-5}\) \\

Claude
& Gemini
& 15.5
& \(1.74 \times 10^{-7}\)
& 0.0
& \(3.41 \times 10^{-8}\)
& 37.5
& \(2.99 \times 10^{-7}\) \\

\bottomrule
\end{tabular}
\end{table*}

\begin{figure*}[t]
    \centering
    \includegraphics[
        width=0.9\textwidth,
        trim={0 0 0 0},
        clip
    ]{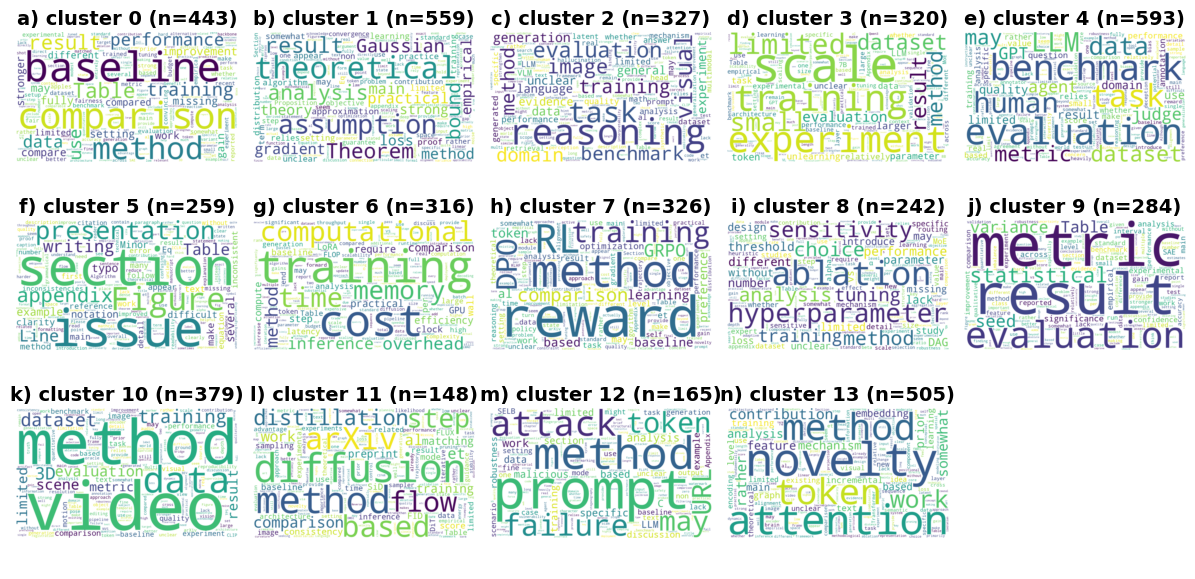}
    \caption{Word clouds for the 14 weakness clusters identified through UMAP dimensionality reduction and \(k\)-means clustering. The value \(n\) indicates the number of weakness statements assigned to each cluster, and larger words represent terms appearing more prominently within the corresponding cluster.}
    \label{fig:weakness_word_clouds}
\end{figure*}

\begin{figure*}[t]
    \centering
    \includegraphics[
        width=0.80\textwidth,
        trim={0 0 0 0},
        clip
    ]{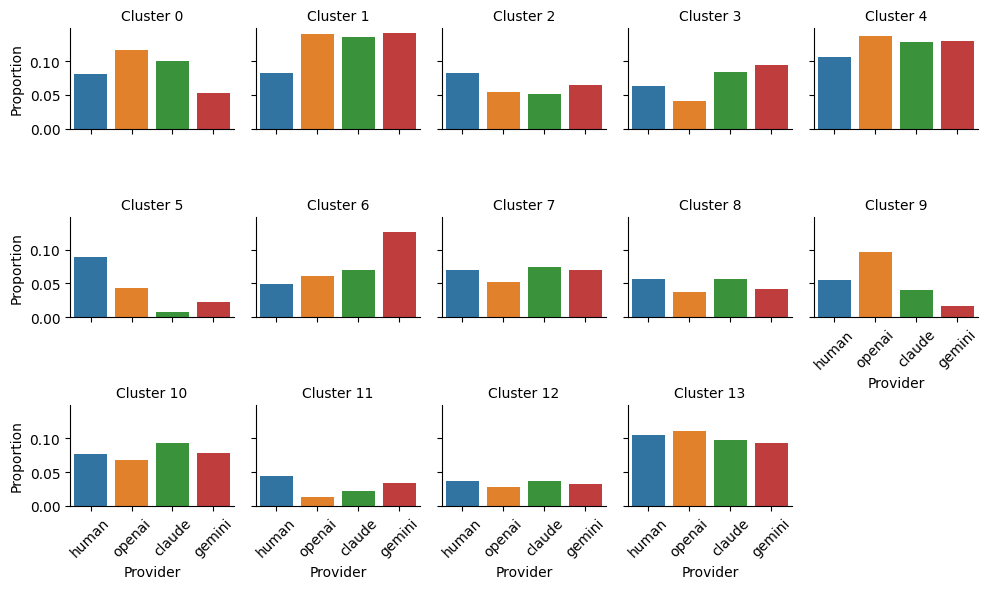}
    \caption{Proportion of weakness statements assigned to each cluster for human, OpenAI, Claude, and Gemini reviewers. Each panel represents one of the 14 weakness clusters.}
    \label{fig:weakness_cluster_distribution}
\end{figure*}


\subsection{Data Collection and Paper Selection}
\label{subsec:data_collection}

We collected ICLR 2026 submission and review data using the OpenReview API. Of the 13,763 submissions reported by ICLR, we retrieved 13,718 papers whose venue metadata was labelled ``Submitted to ICLR 2026.'' For each submission, we collected its OpenReview identifiers, title, abstract, author information, decision, number of reviews, reviewer ratings and confidence scores, and meta-review summary. We also retrieved all available official reviews and their structured evaluation fields. The review-level variables used in the study are summarized in Table~\ref{tab:review_metadata}.

The dataset contained 224 oral papers, 5,128 poster papers, and 8,366 rejected papers. Because the oral category was the smallest, we retained all 224 oral papers and selected 224 thematically similar poster papers and 224 rejected papers.

To match papers by topic, we generated embeddings of their abstracts using OpenAI's \texttt{text-embedding-3-small} model. We calculated cosine similarity and applied one-to-one matching that maximized the total similarity across the full set. Each poster or rejected paper could be matched only once.

From the matched groups, we downloaded 100 papers per category, resulting in a final sample of 300 papers: 100 oral, 100 poster, and 100 rejected papers. We took the average rating given by the human reviewers as the human baseline.

\subsection{LLM Review Generation}
\label{subsec:llm_review_generation}

Each paper was independently evaluated by three models: OpenAI \texttt{gpt-5.4}, Google \texttt{gemini-3.1-pro-preview}, and Anthropic \texttt{claude-opus-4-6}. \textbf{The models were selected because their training cutoffs predated the ICLR 2026 submission period, reducing the risk that they had encountered the evaluated papers during training. Each model generated one review per paper using identical instructions, with external search and browsing disabled to further mitigate data leakage.} The complete prompt is provided in Appendix~\ref{app:review_prompt}.

\subsection{Rating Distribution Analysis}
\label{subsec:rating_analysis}

We first examined whether each LLM differentiated between accepted and rejected papers. Oral and poster papers were combined into an accepted group and compared with the rejected group using a two-sided Mann--Whitney (U) test for each model.

This test was selected because the ratings were discrete, ordinal-like values from two independent groups. It does not require normally distributed data or equal intervals between rating categories \cite{mann1947test,fay2010wilcoxon}. The results were interpreted as differences in the relative rating distributions rather than strictly as differences between medians, since a median-based interpretation requires additional assumptions about the shapes of the distributions \cite{hart2001mann}.

We then compared the three decision categories: Reject, Accept (Poster), and Accept (Oral). For each evaluator, pairwise two-sided Mann--Whitney (U) tests were conducted for Reject versus Poster, Reject versus Oral, and Poster versus Oral. The three comparisons were treated as one family, and their (p)-values were adjusted using the Holm procedure \cite{holm1979simple}.

Because all reviewers evaluated the same papers, comparisons among evaluators used paired two-sided Wilcoxon signed-rank tests matched by submission ID. For RQ3, we compared each pair of LLMs within each decision category. For RQ4, we compared each LLM rating with the paper-level mean human rating. Holm correction was applied to the three comparisons within each category.

\subsection{Weakness Theme Analysis}
\label{subsec:weakness_analysis}

We analyzed the weaknesses identified by human reviewers and the three LLMs through thematic clustering. An embedding was generated for each individual weakness statement.

UMAP was used to reduce the dimensionality of the embeddings. We performed a grid search using component values of 3, 5, 6, 8, and 10 and neighbour values of 10, 15, 20, 25, and 30. Trustworthiness was used to select the configuration that best preserved local relationships from the original embedding space.

The reduced embeddings were clustered using (k)-means. The number of clusters was selected by jointly examining inertia, silhouette score, Calinski--Harabasz index, and Davies--Bouldin index.

Resulting clusters were interpreted using word clouds and by reviewing the weakness statements assigned to each cluster. We also examined the distribution of human-, GPT-, Gemini-, and Claude-generated weaknesses across the clusters to compare the themes emphasized by each reviewer type.

\section{Results}
\label{sec:results}

\subsection{RQ1: Can LLM Scores Distinguish Accepted and Rejected Papers?}
\label{subsec:rq1}

Figure~\ref{fig:accepted_rejected_distributions} presents the score distributions for accepted and rejected papers. Human ratings covered the full scale from 0 to 10, whereas none of the three LLMs used either endpoint. Gemini also assigned a score of 8 to 16 rejected papers, a pattern not observed for OpenAI or Claude.

For accepted papers, the modal score was 6 for OpenAI and Claude and 8 for Gemini. The human modal score was 6. For rejected papers, OpenAI, Claude, and human reviewers had a modal score of 4, whereas Gemini had a modal score of 6.

Table~\ref{tab:accepted_rejected_comparison} shows the Mann--Whitney \(U\) comparisons between accepted and rejected papers within each LLM. All three LLMs assigned significantly higher ratings to accepted papers. The median decreased from 6 to 4 for OpenAI and Claude and from 8 to 6 for Gemini. The differences were significant for Claude (\(p = 1.36 \times 10^{-8}\)), OpenAI (\(p = 4.09 \times 10^{-10}\)), and Gemini (\(p = 2.21 \times 10^{-5}\)). Therefore, all three models distinguished accepted papers from rejected papers at the distribution level.

\subsection{RQ2: Do LLM Scores Track the ICLR Presentation Category?}
\label{subsec:rq2}

Figure~\ref{fig:decision_category_distributions} shows the rating distributions for rejected, poster, and oral papers. Table~\ref{tab:decision_category_comparisons} presents the pairwise comparisons among these categories.

All three LLMs produced significantly different rating distributions for rejected papers compared with both poster and oral papers. However, none distinguished between poster and oral papers. The Holm-adjusted (p)-values for the poster--oral comparison were 0.256 for Claude, 0.376 for OpenAI, and 0.809 for Gemini.

Human ratings showed significant differences between all three decision categories. The median paper-level rating increased from 4.0 for rejected papers to 5.33 for poster papers and 6.5 for oral papers. The difference between poster and oral papers was statistically significant (Holm-adjusted \(p = 3.24 \times 10^{-9}\)). This indicates that the distinction between poster and oral papers was present in the human ratings but was not captured by any of the LLMs.

The results therefore show that the models identified the broader distinction between rejected and accepted papers but did not reproduce the finer distinction between poster and oral presentations.

\subsection{RQ3: Are Different LLMs Calibrated Differently?}
\label{subsec:rq3}

Figure~\ref{fig:llm_rating_comparison} compares the rating distributions of the three LLM providers across the decision categories. Table~\ref{tab:llm_pairwise_comparisons} presents the paired Wilcoxon signed-rank comparisons among the models, with Holm-adjusted $p$-values.

Gemini's ratings differed significantly from those of both OpenAI and Claude for rejected, poster, and oral papers. OpenAI and Claude did not differ significantly for rejected papers (adjusted $p = 1.000$), but their ratings differed significantly for poster papers (adjusted $p = 8.10 \times 10^{-3}$) and oral papers (adjusted $p = 2.65 \times 10^{-2}$).

These findings show that the models followed different scoring patterns. Gemini was particularly distinct because it generally assigned higher ratings, whereas OpenAI and Claude were more similar, especially when evaluating rejected papers.

\subsection{RQ4: Are LLMs More Critical Than Human Reviewers?}
\label{subsec:rq4}

\begin{table}[t]
\centering
\caption{Paired Wilcoxon signed-rank comparisons between human and LLM ratings within each decision category. Wilcoxon \(W\) statistics and Holm-adjusted \(p\)-values are reported.}
\label{tab:llm_human_comparisons}

\scriptsize
\setlength{\tabcolsep}{5pt}
\renewcommand{\arraystretch}{1.15}

\resizebox{\columnwidth}{!}{%
\begin{tabular}{
    @{}
    ll
    cc
    c
    c
    @{}
}
\toprule

\textbf{Decision Category} &
\textbf{LLM} &
\shortstack{\textbf{Human}\\\textbf{Median}} &
\shortstack{\textbf{LLM}\\\textbf{Median}} &
\(\boldsymbol{W}\) &
\shortstack{\textbf{Holm-adjusted}\\\(\boldsymbol{p}\)} \\

\midrule

\multirow{3}{*}{Reject}
& Claude
& 4.0 & 4.0 & 413.5 & \(1.000\) \\

& OpenAI
& 4.0 & 4.0 & 456.0 & \(1.000\) \\

& Gemini
& 4.0 & 6.0 & 68.0 & \(7.44 \times 10^{-7}\) \\

\midrule

\multirow{3}{*}{Accept (Poster)}
& Claude
& 5.33 & 6.0 & 429.0 & \(0.221\) \\

& OpenAI
& 5.33 & 6.0 & 337.5 & \(0.201\) \\

& Gemini
& 5.33 & 8.0 & 36.5 & \(4.09 \times 10^{-7}\) \\

\midrule

\multirow{3}{*}{Accept (Oral)}
& Claude
& 6.5 & 6.0 & 23.5 & \(9.42 \times 10^{-7}\) \\

& OpenAI
& 6.5 & 6.0 & 252.0 & \(0.0325\) \\

& Gemini
& 6.5 & 8.0 & 97.0 & \(4.28 \times 10^{-5}\) \\

\bottomrule
\end{tabular}%
}
\end{table}

Table~\ref{tab:llm_human_comparisons} compares LLM ratings with paper-level human ratings within each decision category using paired Wilcoxon signed-rank tests.

For rejected papers, OpenAI and Claude did not differ significantly from humans, whereas Gemini assigned significantly higher ratings. For poster papers, neither OpenAI nor Claude differed significantly from humans, while Gemini assigned significantly higher ratings. For oral papers, all three LLMs differed significantly from humans: OpenAI and Claude assigned lower ratings, whereas Gemini assigned higher ratings.

Overall, LLMs were not consistently more critical than human reviewers. OpenAI and Claude were more critical primarily when evaluating oral papers, whereas Gemini generally produced more favourable ratings across all three decision categories.

\subsection{RQ5: What Weaknesses Do LLMs Repeatedly Mention?}
\label{subsec:rq5}

The inertia curve did not present a clear elbow for selecting the number of clusters. We therefore considered the silhouette and Davies--Bouldin results, selecting a local maximum in the silhouette score that corresponded with a local minimum in the Davies--Bouldin index. Based on these results, 14 weakness clusters were selected.

The identified themes were as follows:

\begin{enumerate}
\item Weak or insufficient experimental comparison with baselines.
\item Limitations and weaknesses in the theoretical analysis.
\item Limited experimental scope and insufficient evidence for broad reasoning or generalization claims.
\item Limited scalability and generalizability due to small-scale experimental validation.
\item Questionable reliability and validity of benchmarks, datasets, and evaluation metrics.
\item Poor writing, presentation, and technical clarity.
\item Insufficient analysis of computational efficiency and resource costs.
\item Insufficient justification and validation of reinforcement-learning and reward-design choices.
\item Insufficient ablation studies and hyperparameter sensitivity analysis.
\item Insufficient statistical rigor and incomplete analysis of experimental results.
\item Practical limitations and restricted applicability of the proposed method.
\item Limited novelty and inadequate comparison with closely related diffusion and flow-based methods.
\item Incomplete security evaluation and limited robustness against realistic or adaptive attacks.
\item Limited methodological novelty and insufficient justification of the proposed design.
\end{enumerate}

Figure~\ref{fig:weakness_word_clouds} presents the word clouds used to support the interpretation of the clusters. Figure~\ref{fig:weakness_cluster_distribution} shows the distribution of weaknesses generated by human reviewers and the three LLMs. Clusters 1, 4, and 13 were among the most frequent themes overall. Weak experimental comparison with baselines, represented by Cluster~0, was particularly common in the LLM reviews. In contrast, insufficient analysis of computational efficiency and resource costs, represented by Cluster~6, was particularly common in the human reviews.

These results suggest that human and LLM reviewers shared several major criticism themes but differed in the relative attention given to specific weaknesses.

\subsection{RQ6: Where Do LLMs Disagree Most with Humans?}
\label{subsec:rq6}

The largest disagreements between LLMs and human reviewers appeared in the assessment of paper quality and in the types of weaknesses emphasized. Human ratings significantly distinguished oral from poster papers, whereas none of the LLMs captured this difference. Gemini consistently assigned higher ratings than humans, including a median rating of 6 for rejected papers and a rating of 8 for 16 rejected papers. In contrast, OpenAI and Claude assigned lower median ratings than humans for oral papers.

LLMs also used a narrower portion of the rating scale, as none assigned ratings of 0 or 10, both of which were used by human reviewers. The weakness analysis showed different priorities: LLMs more frequently criticized insufficient baseline comparisons, whereas human reviewers more often raised concerns about computational efficiency and resource costs.

Overall, the strongest disagreements occurred when evaluating oral papers and when determining which weaknesses were most important.

\section{Conclusions}
\label{sec:conclusion}

This study compared peer reviews generated by three contemporary LLMs with human reviews and final decisions for 300 topic-matched ICLR 2026 submissions. We examined whether LLM ratings aligned with broad acceptance outcomes and finer presentation categories, whether providers used the recommendation scale similarly, and whether human and LLM reviewers emphasized comparable weaknesses.

All three LLMs assigned higher ratings to accepted papers than to rejected papers, indicating that their scores captured a broad signal associated with conference acceptance. However, none reproduced the distinction between poster and oral papers observed in the human ratings. This finding suggests that broad accept--reject discrimination should not be interpreted as evidence that an automated reviewer captures finer differences reflected in conference decisions.

The models also exhibited distinct scoring patterns. Gemini generally assigned more favorable ratings, including comparatively high scores to some rejected papers. OpenAI and Claude were more similar to human reviewers for rejected and poster papers but assigned comparatively lower ratings to oral papers. In addition, the LLMs used a narrower portion of the recommendation scale than human reviewers. These differences demonstrate that conclusions about automated reviewing may depend substantially on the selected model.

The analysis of written weaknesses revealed both shared themes and systematic differences in emphasis. Human and LLM reviewers frequently identified concerns related to experimental evidence, theoretical analysis, generalizability, clarity, and methodological novelty. However, LLMs more often criticized insufficient comparisons with baselines, whereas human reviewers more frequently emphasized computational efficiency and resource requirements. Alignment in numerical ratings therefore does not necessarily imply alignment in the concerns considered most important.

Overall, our findings support a limited but potentially useful role for LLMs in scientific reviewing. They may provide supplementary feedback, identify recurring weaknesses, and help authors examine their work from multiple perspectives. Their ratings and priorities, however, should not be treated as interchangeable with human peer-review judgments. Future automated-review systems should be evaluated not only by the fluency of their reviews or their ability to distinguish accepted from rejected papers, but also by their stability, scale usage, evidence grounding, and alignment with the finer structure of expert evaluation.

\paragraph{Limitations.}
\label{sec:limitations}

This study has several limitations. The analysis is based on 300 topic-matched submissions from a single conference year and may not represent the full distribution of ICLR papers or generalize to other venues, years, or research communities. Final decisions and presentation categories are also imperfect proxies for scientific quality because they reflect reviewer judgments, discussion, capacity constraints, and conference-specific criteria; therefore, the inability of the LLMs to distinguish oral from poster papers should be interpreted as a lack of alignment with this decision structure rather than definitive evidence that they cannot recognize scientific excellence. Comparisons with human reviewers are further complicated by differences in reviewer expertise, confidence, severity, and the number of reviews per paper.




\appendix

\section{LLM Review Prompt}
\label{app:review_prompt}

The following prompt was provided to all three models:

\begin{quote}
\small

Take the role of a reviewer for ICLR 2026.

The following recommended step-by-step process for a normal ICLR reviewer is provided as general context for evaluating the paper. You do not need to follow each step explicitly or reproduce the process in your response. Use it only to guide the overall criteria and reasoning behind your assessment.
\begin{itshape}
\begin{enumerate}
    \item Read the paper: It is important to carefully read through the entire paper and to look up any related work and citations that will help you comprehensively evaluate it. Be sure to give yourself sufficient time for this step.

    \item While reading, consider the following:
    \begin{enumerate}
        \item Objective of the work: What is the goal of the paper? Is it to better address a known application or problem, draw attention to a new application or problem, or introduce and/or explain a new theoretical finding? A combination of these? Different objectives will require different considerations regarding potential value and impact.

        \item Strong points: Is the submission clear, technically correct, experimentally rigorous, and reproducible? Does it present novel findings, for example, theoretically or algorithmically?

        \item Weak points: Is it weak in any of the aspects listed above?

        \item Be mindful of potential biases and try to be open-minded about the value and interest a paper can hold for the entire ICLR community, even if it may not be very interesting to you.
    \end{enumerate}

    \item Answer four key questions for yourself to make a recommendation to Accept or Reject:
    \begin{enumerate}
        \item What is the specific question and/or problem tackled by the paper?

        \item Is the approach well motivated, including being well placed in the literature?

        \item Does the paper support its claims? This includes determining whether the results, theoretical or empirical, are correct and scientifically rigorous.

        \item What is the significance of the work? Does it contribute new knowledge and sufficient value to the community? This does not necessarily require state-of-the-art results. Submissions bring value to the ICLR community when they convincingly demonstrate new, relevant, and impactful knowledge, including empirical, theoretical, or practitioner-oriented knowledge.
    \end{enumerate}

    \item Write and submit your initial review, organizing it as follows:
    \begin{enumerate}
        \item Summarize what the paper claims to contribute. Be positive and constructive.

        \item List the strong and weak points of the paper. Be as comprehensive as possible.

        \item Clearly state your initial recommendation, Accept or Reject, with one or two key reasons for this choice.

        \item Provide supporting arguments for your recommendation.

        \item Ask questions that you would like the authors to answer to clarify your understanding of the paper and provide the additional evidence needed to be confident in your assessment.

        \item Provide additional feedback aimed at improving the paper. Make it clear that these points are intended to help and are not necessarily part of your decision assessment.
    \end{enumerate}
\end{enumerate}
\end{itshape}

Return your response as a valid JSON object with the following fields:

\begin{verbatim}
{
  "summary": "A concise summary of what the 
  paper claims to contribute.",
  "strengths": [
    "One or more strengths of the paper."
  ],
  "weaknesses": [
    "One or more weaknesses of the paper."
  ],
  "questions": [
    "One or more questions for the authors."
  ],
  "rating": value,
  "confidence": value
}
\end{verbatim}

For \texttt{"rating"}, return only one of the following numbers:

\begin{itemize}
    \item 0 = strong reject
    \item 2 = reject, not good enough
    \item 4 = marginally below the acceptance threshold
    \item 6 = marginally above the acceptance threshold
    \item 8 = accept, good paper
    \item 10 = strong accept, should be highlighted at the conference
\end{itemize}

For \texttt{"confidence"}, return only one of the following numbers:

\begin{itemize}
    \item 1 = You are unable to assess this paper and have alerted the ACs to seek an opinion from different reviewers.

    \item 2 = You are willing to defend your assessment, but it is quite likely that you did not understand the central parts of the submission or that you are unfamiliar with some pieces of related work. Math or other details were not carefully checked.

    \item 3 = You are fairly confident in your assessment. It is possible that you did not understand some parts of the submission or that you are unfamiliar with some pieces of related work. Math or other details were not carefully checked.

    \item 4 = You are confident in your assessment, but not absolutely certain. It is unlikely, but not impossible, that you did not understand some parts of the submission or that you are unfamiliar with some pieces of related work.

    \item 5 = You are absolutely certain about your assessment. You are very familiar with the related work and checked the math or other details carefully.
\end{itemize}

The output must be valid JSON only. Do not include any text outside the JSON object.

\end{quote}

\section{Computational Cost}

All experiments were run on a single laptop and required approximately three days of wall-clock time. The total API cost was approximately 300 USD, consisting of about 100 USD in OpenAI usage 100 USD in Gemini usage, and 100 USD in Claude usage.

\section{Use of AI Assistants}

Beyond their use in the experimental pipelines, LLMs were also used as writing and research assistants during the preparation of the paper, including for editing, contrasting ideas, and supporting literature search.

\bibliography{custom}

\end{document}